# Analysis and mining of low-carbon and energy-saving tourism data characteristics based on machine learning algorithm

Lukasz Wierzbinski

ICAIE Organizing Committee, Poland

**Abstract:** In order to study the formation mechanism of residents' low-carbon awareness and provide an important basis for traffic managers to guide urban residents to choose low-carbon travel mode, this paper proposes a low-carbon energy-saving travel data feature analysis and mining based on machine learning algorithm. This paper uses data mining technology to analyze the data of low-carbon travel questionnaire, and regards the 15-dimensional problem under the framework of planned behavior theory as the internal cause variable that characterizes residents' low-carbon travel willingness. The author uses K-means clustering algorithm to classify the intensity of residents' low-carbon travel willingness, and applies the results as the explanatory variables to the random forest model to explore the mechanism of residents' social attribute characteristics, travel characteristics, etc. on their low-carbon travel willingness. The experimental results show that based on the Silhouette index test and t-SNE dimensionality reduction, residents' low-carbon travel willingness can be divided into three categories: strong, neutral, and not strong; Based on the importance index, the four most significant factors are the occupation, residence, family composition and commuting time of residents. Conclusion: This method provides policy recommendations for the development and management of urban traffic low-carbon from multiple perspectives.

## 1 Introduction

The United Nations Environment Programme (UNPE) reported on March 16, 2008 that glaciers are melting at the fastest rate due to global climate change, and many glaciers may melt within decades. According to the survey of scientists, nearly 30 glaciers around the world retreated by an average of 0.3 m per year from 1980 to 1999; However, since 2000, the retrogression speed has increased to an average of 0.37m per year; The average retreat in 2006 was 1.5m. At the same time, from 1961 to 2003, the annual rise rate of sea level was 1.5 mm, that is to say, the sea level rose about 6.35 cm2 in these 42 years.As far back as 1896, Nobel Prize in Chemistry winner Svante Arrhenius predicted that the burning of fossil fuels would release carbon monoxide $CO_2$ into the atmosphere and cause global warming. Carbon dioxide ($CO_2$) is the main greenhouse gas responsible for climate change. Scientists study the composition of past air by measuring bubbles in Antarctic glaciers. Research shows that the content of $CO_2$ in the air is the highest in the past 420000 years. The $CO_2$ content in the air has stabilized at 280 ppm in the past 1000 years, but it has increased sharply to 380 ppm by the end of the 20th century. According to the different scenarios outlined by the IPCC, it is predicted that the content of $CO_2$ in the air will rise further to 550 ppm to 960 ppm by 2100 [1]. Since the United Nations General Assembly decided to launch the negotiation of the international climate convention in 1990 and the Kyoto Protocol came into force in 2005, the Bali Road Map in 2007 and the United Nations Climate Change Conference in Copenhagen at the end of 2009 have all reflected the consensus reached by the international community to reduce C02 emissions.

As China's urbanization process progresses, the total demand for urban traffic from urban residents increases, which is not only convenient for large urban traffic, but also poses a threat to urban air. The tourism behavior of residents is an important source of energy consumption, and its impact on the carbon emissions of the whole society is clear [2]. In order to reduce carbon emissions and encourage residents to choose greener, lower-carbon travel models, there is a need to quickly pay attention to the impact of environmental and environmental behaviors and other aspects of behavior in addition to innovations in transportation technology. Therefore, it is theoretically important and important to find out the facts that affect the satisfaction of urban residents to guide urban residents to choose low-carbon, reduce carbon emissions, and build low-carbon cities. Those. Figure 1 shows a survey of the low-carbon and energy-saving construction industry.

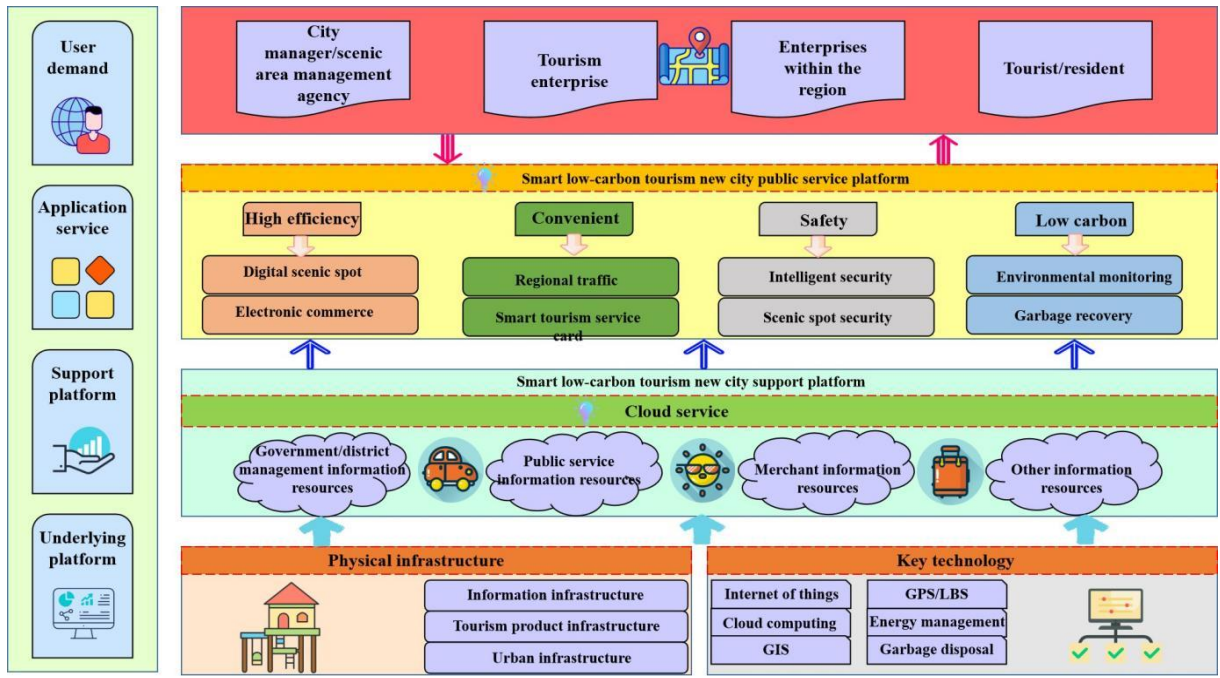

Figure 1 Research on the development framework of low-carbon and energy-saving tourism

## 2 Literature review

Dong, K. et al. Using panel data from 30 provinces in China between 2004 and 2007, a non-parametric panel quantile causality (PCIQ) method was used to assess the relationship between less electricity and energy poverty. An important finding of our study is that there is a positive relationship between low energy consumption and reduced poverty. Small-scale energy transitions can help reduce energy poverty by affecting access to energy services, clean energy, energy management, availability energy and performance [3]. Khan, S. et al examine the impact of low carbon, crime and transport on tourism and trade in the Association of Southeast Asian Nations (ASEAN) by 2011-2019 balance sheet. The model is prepared with four 6 hypotheses. the model shows that transportation and energy use have a positive effect on tourism, while crime in ASEAN countries affects the economy. As an ASEAN economy, it has a positive impact on transportation, renewable energy, and tourism. Therefore, it can be concluded that improving the ecology and reducing carbon dioxide emissions through the development of renewable energy, good logistics services and infrastructure, and creating peace conceived with less crime will encourage the arrival of tourists and have a positive effect on the economy of ASEAN countries[4]. Coniglio, S. et al. propose a framework to solve these problems, which uses different factors to predict the probability of adverse events before they occur. In particular, the focus is on two low-carbon applications: foaming during anaerobic digestion and condenser tube loss in nuclear power plant steam turbines. A number of experiments clearly show the effectiveness of our technology [5].

This study is based on the existing low carbon travel questionnaire data of residents (including individual travel information data and travel attitude data), and combines K-means clustering and random forest algorithm to mine the survey data; We classify the intensity of residents' low-carbon travel intention, and discuss the mechanism of residents' personal attributes, travel characteristics, etc. on the formation of residents' low-carbon travel intention; The research results provide policy recommendations for the low-carbon development and management of urban traffic from multiple perspectives.

## 3 Research methods

### 3.1 Data collection and analysis of low-carbon travel intention

This article uses a "question star" to collect data on the carbon monoxide journey. According to the concept of behavior planning technique, the behavior of the research participants was investigated by the five items of individual content, behavioral control, behavioral personality, psychological needs, and behavior.The first four were cause variables, and the last one was outcome variables, as shown in Table 1.

Table 1 Low-carbon travel willingness questionnaire

| Variable category | Observation variable serial number | Item |
|---|---|---|
| Behavior and attitude | 1 | I think it is necessary to reduce the traffic in order to solve the traffic problem and improve the urban environment. |
| | 2 | I think that choosing a low-carbon type can solve the problem of environmental pollution more than other types of travel. |
| | 3 | I am willing to choose a low-carbon model to achieve my daily tasks |
| Subjective norm | 4 | People around me are used to using lower carbon travel mode |
| | 5 | People around me want me to choose low-carbon models to get to my daily work. |
| | 6 | Increased carbon monoxide travel has a positive effect on my travel mode choices |
| Perceptual behavior control | 7 | Advances in public bicycles make me prefer low-carbon travel |
| | 8 | The development of public transport makes me more inclined to travel by public transport |

| | 9 | The increase in taxi fares has led to the desire to choose other low-carbon travel methods |
|---|---|---|
| | 10 | Shopping restrictions and the cost of using a car made me want to choose other low-carbon travel methods |
| Common psychological needs | 11 | I usually choose the travel mode that can reach the destination as soon as possible |
| | 12 | I usually choose a more comfortable mode of transportation |
| | 13 | If I can save time, I am willing to spend more money |
| Behavioral intention | 14 | I plan to choose a lower carbon travel mode in most travel |
| | 15 | I will probably choose a lower carbon travel mode |

All the indicators of this questionnaire are evaluated using a five-level Likert rating model, where "same, same, general, not uniform, not uniform" correspond to 1, 2, 3, 4, and 5 points. Those. In addition, we collected information such as gender, age, education level, monthly personal income; A total of 2941 valid studies were excluded from the database. The results of the survey show that, from a behavioral perspective (corresponding to analysis variables 1, 2, and 3), when the number of strong supporters is above 30, the number of strong supporters is not. strong; From the point of view of simple ideas (included changes 4, 5, and 6 of the analysis), only 10 of the people have something in common, while almost 45 have half of the things use, and almost 12 nothing in most cases. ; For the concept of behavioral control (such as identifying differences 7, 8, 9, and 10), the number of people who agree more than 15, the number of people who see good with the average personality is 38, and people. . Disagree almost 10; From the point of view of mental needs (screening changes 11, 12, 13), almost 18 are the same, and people do not agree that "if I can save time, I want to spend more ." time more than money" is more than people agree; Money "Behavior, about 16 agree, 44 agree, 34 neutral, and 6 disagree [6-8].

### 3.2 K-means clustering algorithm for low-carbon travel intention data

The questionnaire data meets the consistency test, that is, the individual's behavioral incentives and behavioral results are matched. We can understand that individuals with similar behavioral intentions (corresponding to observation variables 14 and 15) have similar behavioral incentives, that is, the selection results of cause variables (observation variables 1-13) in the questionnaire are similar. That is to say, the intensity of individual low carbon travel willingness reflected by individual subjective norms, perceptual behavior control and other subconscious is consistent. The paper does not discuss how individuals form low-carbon travel intentions, but focuses on the low-carbon travel intentions of individuals, which are reflected by 15 observation variables, and classifies groups according to the intensity of the intentions. For this research, there is no relevant literature. This research method makes full use of the questionnaire data, which will make up for the defect of relying solely on the survey results of "behavioral intention" in the questionnaire to judge residents' travel intention, and make the identification results of residents' behavioral intention more reliable. On the other hand, we can analyze the differences of the attribute characteristics of different types of people and fill the gap in the current research [9].

MacQueen proposed K-means clustering algorithm for the first time, which belongs to unsupervised learning hard clustering algorithm and searches for the best clustering through iteration. One of the core problems of clustering is how to define the similarity measure function reasonably. Different from the common quantitative data similarity measurement, the data in this paper are qualitative data, so the sequence alignment method is introduced here.

### 3.2.1 Definition of similarity function based on sequence alignment

In bioinformatics, sequence alignment is a basic research method. Sequence similarity can determine the homology relationship between sequences. Here, the questionnaire consists of 15 questions (as shown in Table 1), each of which has 5 options (Likert's five-level scoring method). Each respondent's selection results for these 15 questions form a string in form, such as "1, 2, 3,..., 5, 2, 3", which is similar to the nucleic acid or protein sequence in biology. Here, we use the sequence data processing method for reference, and measure the heterogeneity of people's willingness to low-carbon travel based on the similarity between strings. Similar data processing methods are also applied to traffic accident analysis and stock prediction. The indicators to measure the similarity of two strings are Levenshtein distance ($L_{AB}$) and European distance ($D_{AB}$).

1) Levenshtein distance ($L_{AB}$)

Levenshtein distance is the editing distance, which is a concept put forward by Vladimir Levenshtein in 1965. It refers to the minimum number of editing operations required to convert one string to another between two strings. The smaller the editing distance, the more similar the two strings are. When the editing distance is 0, the two character strings are equal. Suppose that the two strings are s1 and s2, and their length is $m$ and $n$, first set a matrix of $(m + 1)(n + 1)$ size, and then initialize the first row and the first column: $d[i, 0] = i, d[0, j] = j$, and calculate the editing distance through formula (1):

$$d_{ij} = \begin{cases} 0, i = 0 \text{ or } j = 0 \\ min(d_{[i-1,j]} + 1, d_{[i,j-1]} + 1, d_{[i-1,j-1]}), \\ x_i = y_i \\ min(d_{[i-1,j]} + 1, d_{[i,j-1]} + 1, d_{[i-1,j-1]} + \\ 1), x_i \neq y_i \end{cases} \quad (1)$$

2) European distance ($D_{AB}$)

The editing distance can reflect the similarity in the composition of two strings, but it cannot reflect the "distance" between two strings, that is, consistency [10]. In this paper, the Likert five-level scoring method is used. There is a gap between the people who choose "1" and the people who choose "5" or "2" in a certain variable. In order to reflect the differences in people's attitudes, we can use distance indicators to measure. Here, Euclidean distance, which is widely used, is used to measure the distance between samples. The distance between points $x = (x_1, \ldots, x_i)$ and $y = (y_1, \ldots, y_i)$ is shown in formula (2):

$$d(x,y) = \sqrt{\sum_{i=1}^{n} (x_i - y_i)^2} \quad (2)$$

In order to reflect the similarity and consistency of the two strings, the paper combines the advantages of editing distance and European distance to build a comprehensive index $S_{AB}$ for similarity measurement of the will data, as shown in formula (3):

$$S_{AB} = (1 + L_{AB})D_{AB} \quad (3)$$

### 3.2.2 Implementation of K-means clustering algorithm for low-carbon travel intention

The implementation steps of the K-means clustering algorithm based on similarity measurement are as follows:

Step 1 In this paper, for a given dataset $\{x^{(1)}, x^{(2)}, \ldots, x^{(15)}\}$ with 15-dimensional data, randomly select k initial cluster centroids $\{\mu_1, \mu_2, \ldots, \mu_k\}$.

Step 2 For each data $x^{(i)}$, calculate its category as follows (4):

$$c^{(i)} = \arg \min_{j} \|x^{(i)} - \mu_j\|^2 \quad (4)$$

For each type of $j$, recalculate the centroid of this type as follows (5):

$$\mu_j = \frac{\sum_{i=1}^{m} l\{c^{(i)} = j\}x^{(i)}}{\sum_{i=1}^{m} l\{c^{(i)} = j\}} \quad (5)$$

Where, $k$ represents the number of clusters given in advance, and $c^{(i)}$ represents the closest class between data $x^{(i)}$ and $k$ classes.

Step 3 Repeat Step 2 until convergence.

# 4 Result analysis

## 4.1 K-means clustering analysis of low-carbon travel intention

In this paper, the algorithm is implemented with the help of Matlab. According to the selection results of residents' low-carbon travel willingness, the group can be roughly divided into 2 categories (strong, not strong) or 3 categories (strong, neutral, not strong). The commonly used indicators to test the effectiveness of clustering include Calinski-Harabase (CH), Weightedinter-intra (Winter), In-Group Promotion (IGP) and Silhouette, among which Silhouette is widely used for its simplicity and good evaluation ability. The best classification results are judged according to Silhouette. The value of the Silhouette Performance index is changed to [- 1, 1]. The largest is the average Silhouette Performance Index value of each model, the largest is the clustering zoo. The number of pagw sib haum rau tus nqi the highest is the highest pagw zoo [11-12]. This is shown in Figure 2 from the Silhouette measurement of 2 groups and 3 groups.

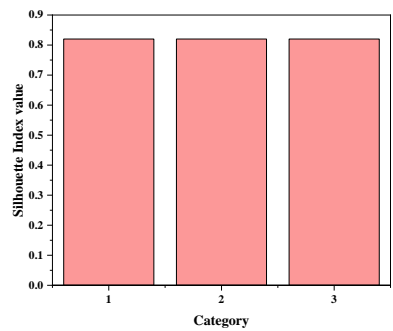

(a) 3 categories

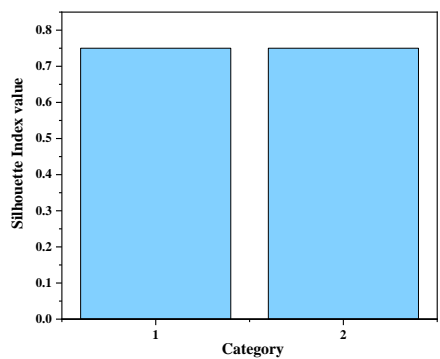

(b) 2 categories

Figure 2 Silhouette Index Value

It can be seen from Figure 2 that the average Silhouette index value of samples divided into three categories (strong, neutral, and not strong) is above 0.8, which is significantly greater than the result of samples divided into two categories (strong, not strong); Therefore, it is better to divide the sample into 3 categories than 2 categories. The multi-dimensional centroid value obtained based on K-means clustering is shown in Figure 3.

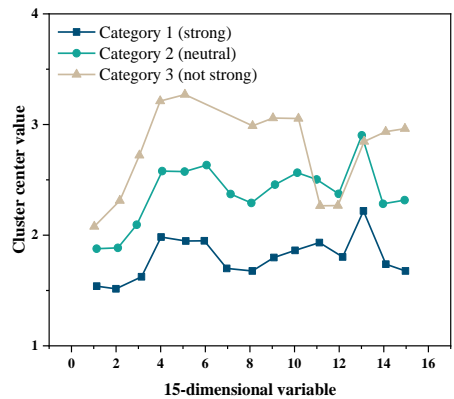

Figure 3 K-means clustering centroid distribution

From the distribution of Figure 3, among the three categories, the average score of category 1 is higher, which is a group with strong low-carbon travel intention; The group with the lowest average score is category 3, which is the group with low carbon travel intention. By analogy, category 2 in the middle is a group with low carbon travel willingness.

## 4.2 Analysis on the formation mechanism of residents' low-carbon travel willingness based on random forest

People who engage in low-carbon travel are influenced by many factors, such as age, gender, education level, travel characteristics such as travel type, distance, and time, as well as their social location, related policies, and systemic economic issues. The distribution of data from this study among low-carbon residents is complex, multi-trait, large noise, and has many factors that determine behavioral variation, making it difficult to capture the model of interest estimated by traditional methods. Because random forests have good width and noise, it is necessary to model human effects on carbon monoxide travel. A random forest is a supervised learning distribution system. For multivariate data, random forest performance indicators (classification accuracy, optimization algorithm, etc.) are better than other classification methods, and group. Therefore, random forest technology has been increasingly used in bioinformatics, text mining, image classification and other fields in recent years, and has become a research hot spot in data mining, machine learning, pattern recognition and other fields [13-14] .

### 4.2.1 Variable selection

The intensity of residents' low carbon travel willingness obtained by K-means clustering is taken as the explanatory variable. Referring to the existing research results, 13 variables reflecting residents' personal characteristics, travel characteristics, etc. are selected as the explanatory variables, among which the socio-economic conditions and relevant policies and systems are inferred according to the residence of residents. See Table 2 for variable names and specific meanings.

Table 2 Classification of influence factors

| Serial number | Variable name | Symbol | Variable |
|---|---|---|---|
| 1 | Low-carbon travel willingness | will | 1 - strong, 2 - neutral, 3 - not strong |
| 2 | Gender | gender | 1-male, 2-female |
| 3 | Age | age | 1- Under18 years old, 2- 18-25 years old, 3- 26~35 years old, 4- 36~45 years old, 5- 46~55 years old, 6 - 55 years old and above |
| 4 | education | education | 1 - junior high school and below, 2 - senior high school, 3 - technical secondary school, 4 - junior college, 5 - undergraduate, 6 - postgraduate, 7 - doctoral and above |
| 5 | Place of residence | address | 1 - Zhenjiang, 2 - Changzhou, 3 - Wuxi, 4 - Suzhou, 5 - Shanghai, 6 - Others |
| 6 | Occupation | career | 1 - full-time students, 2 - production personnel, 3 - sales personnel, 4 - marketing/public relations personnel, 5 - administrative/logistics personnel, 6 - human resources, 7 - financial audit, 8 - civil affairs, 9 - technical research and development, 10 - management personnel, 11 - teachers, 12 - consultants, 13 - professionals (lawyers, architects, medical care, journalists, etc.), 14 - others |
| 7 | Monthly income | income | 1- Less than 2000 yuan, 2- 2001~4000 yuan, 3- 4 001~6000 yuan, 4- 6001~8000 yuan, 5- more than 8000 yuan |
| 8 | Own | media | 1 - bicycle, 2 - electric vehicle, 3 - car, 4 - unit bus, 5 - others |

| | | | vehicles |
|---|---|---|---|
| 9 | Cards and certificates owned | card | 1 - bus card, 2 - public bicycle card, 3 - car driver's license, 4 - other driver's licenses |
| 10 | Family composition (children) | kids | 1 - preschool children, 2 - primary school students, 3 - junior high school students, 4 - senior high school and above, 5 - no |
| 11 | Commuting mode | mode | 1 - walking, bicycle, 2 - electric vehicle, 3 - motorcycle, 4 - taxi, 5 - bus, unit bus or subway, 6 - private car |
| 12 | Commute Distance | distance | 1- 0~5 km, 2-6~10 km, 3-11~15 km, 4-16~20 km, more than 5- 20 km |
| 13 | Commuting time | time | 1- Below 15 min, 2- 16~30 min, 3- 31~45 min, 4- 46~60 min, 5- 61~75 min |

### 4.2.2 Model parameter estimation

In this paper, the randomForest software package of R software is used to build a random forest model of low-carbon travel intention. First, you need to set the initial value of the number of sample predictors mtry at each node, usually the root mean square of the number of variables; Based on the minimum OOB (Out of Bag) of out-of-bag data error, determine the mtry value and the optimal leaf node tree ntree. In the paper, mtry=4, ntree=500. The results show that the classification error of the model is small, in which the error of category 1 is 12.4%, the error of category 2 is 8.3%, and the error of category 3 is 6.7% [15-17].

Enter the value command (rf) to see the value of each variable. MDecA, Mean Reduced Exposure, shows the increase in this variable for the prediction. The bigger the size, the more important the impact of this difference. The order of importance of each factor is shown in Table 3.

Table 3 Ranking of importance of each factor

| Influence factor | MDecA | Influence factor | MDecA |
|---|---|---|---|
| Occupation | 64.7 | Cards and certificates owned | 55.1 |
| Place of residence | 63.6 | Age | 53.6 |
| Family composition | 61.5 | Education | 42.4 |
| Commuting time | 60.1 | Monthly income | 35.4 |

| Commute Distance | 58.3 | Commuting mode | 33.5 |
| Own vehicles | 56.4 | Gender | 32.1 |

As shown in Table 3, from the point of view of characteristics, four factors such as the person's occupation, place of residence, family structure, and working hours; The most important factors are gender, monthly income and type of travel. Taking the most important work as an example, the relationship between work and the purpose of travel is analyzed in the form of a histogram, and the results are shown in Figure 4.

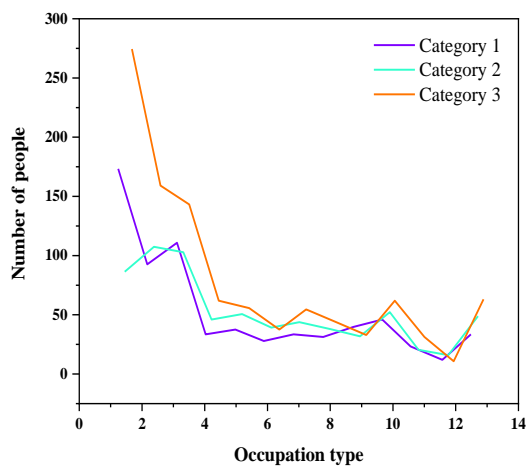

Figure 4 Corresponding relationship between occupation and K-means clustering results

### 4.2.3 Analysis of random forest model results

According to the distribution of jobs in Table 2, except for the 14th category of jobs, from the remaining 13 jobs, according to statistical data, the products of day students and workers are less carbon monoxide. It can be seen from Figure 4 that approximately 50% of the population involved in these two types of work fall into category 3 of carbon monoxide. This is because full-time students and staff are often scheduled to travel to and from school/classroom, and students participating in this program are often high school students, often have a stable and unsocial approach [18].Workplaces Similarly, compared to other workplaces, manufacturing workers have longer daily hours, more on-duty time, and limited options.

From the perspective of family composition, the group with pre-school children in the family said it was not very strong. It is possible that a large number of families also have to bear the problem of transportation to and from school. Therefore, due to the restriction of children's time to and from school, the flexibility of travel mode is reduced.

From the perspective of commuting time, when the commuting time is less than 15min, 70% of the population have a strong desire for low-carbon travel; At 6 to 30 minutes, about 60% of the population have a strong desire for low-carbon travel; At 31~45 min, 40% of people are willing to travel in low carbon. When the commuting time is more than 45 minutes, the number of people with strong desire to travel in low carbon significantly decreases [19-20].

**5 Conclusion**

The survey of residents' low-carbon travel willingness usually includes two aspects of data: external variables such as personal attributes and travel characteristics, and internal variables such as individual attitudes and subjective norms. This study fully exploits the data obtained from the questionnaire survey of low-carbon travel willingness, and analyzes the formation mechanism of residents' low-carbon travel willingness by integrating K-means clustering method and random forest algorithm. The main conclusions of the study are as follows:

(1) The questionnaire is designed based on the planned behavior theory. The questionnaire takes 15 questions as a 15-dimensional characteristic variable to represent residents' low-carbon travel willingness, applies sequence alignment method to similarity measurement, and classifies residents according to the intensity of residents' low-carbon travel willingness based on K-means clustering algorithm; We optimize the number of clusters based on the Silhouette index, and reduce the dimension of the clustering results based on the t-SNE algorithm to verify the validity of the clustering results; Based on the centroid distribution of 15-dimensional data, we extract three types of residents' low-carbon travel willingness, which are strong, neutral and not strong.

(2) As explanatory variables, and uses less carbon monoxide travel game based on different stories. Using the random forest method, the author evaluates the external cause process of carbon monoxide production emissions and the level of related features according to the minor and critical scales; From a characteristic point of view, four factors: self-employment, place of residence, family structure, and travel time are the most important, while female, gender, monthly income, and type of travel are the least important.

(3) Based on the existing questionnaire data, this study is greatly constrained in the selection of internal and external factors, especially in the application process of stochastic forest model, the explanatory variables are not sufficient; In addition, in the design of attitudes, subjective norms and other issues, we have no mature scale for reference, so the survey results will inevitably be affected, and the clustering results will inevitably have deviations. In the follow-up research, we will focus on improving the above problems.

**Reference**

[1]   Jinhong, C., Shuxiao, L., Zheng, W., & Zhanhong, C. (2022). The Characteristic Differences between Ecological Culture and Low-carbon Tourism Cognition under the Vision of Carbon Neutrality. Journal of Resources and Ecology, 13(5), 936-945.

[2]   Tian, X., Zhang, Q., Chi, Y., & Cheng, Y. (2021). Purchase willingness of new energy vehicles: a case study in Jinan City of China. Regional Sustainability, 2(1), 12-22.

[3]   Dong, K. , Ren, X. , & Zhao, J. . (2021). How does low-carbon energy transition alleviate energy poverty in china? a nonparametric panel causality analysis. Energy Economics, 103.

[4]   Khan, S. , Godil, D. I. , Abbas, F. , Shamim, M. A. , & Zhang, Y. . (2022). Adoption of renewable energy sources, low-carbon initiatives, and advanced logistical infrastructure—an step toward integrated global progress. Sustainable Development, 30(1), 275-288.

[5]   Coniglio, S. , Dunn, A. J. , & Zemkoho, A. B. . (2021). Infrequent adverse event prediction in low-carbon energy production using machine learning, 1(2), 29-38.

[6]   Haoxiang, W., & Smys, S. (2021). Big data analysis and perturbation using data mining algorithm. Journal of Soft Computing Paradigm (JSCP), 3(01), 19-28.

[7]   Edastama, P., Dudhat, A., & Maulani, G. (2021). Use of Data Warehouse and Data Mining for Academic Data: A Case Study at a National University. International Journal of Cyber and IT Service Management, 1(2), 206-215.

[8]   Ageed, Z. S., Zeebaree, S. R., Sadeeq, M. M., Kak, S. F., Rashid, Z. N., Salih, A. A., & Abdullah, W. M. (2021). A survey of data mining implementation in smart city applications. Qubahan Academic Journal, 1(2), 91-99.

[9]   Abdullah, D., Susilo, S., Ahmar, A. S., Rusli, R., & Hidayat, R. (2022). The application of K-means clustering for province clustering in Indonesia of the risk of the COVID-19 pandemic based on COVID-19 data. Quality & Quantity, 56(3), 1283-1291.

[10] Chowdhury, K., Chaudhuri, D., & Pal, A. K. (2021). An entropy-based initialization method of K-means clustering on the optimal number of clusters. Neural Computing and Applications, 33(12), 6965-6982.

[11] Stemmer, U. (2021). Locally private k-means clustering. The Journal of Machine Learning Research, 22(1), 7964-7993.

[12] Zhang, E., Li, H., Huang, Y., Hong, S., Zhao, L., & Ji, C. (2022). Practical multi-party private collaborative k-means clustering. Neurocomputing, 467, 256-265.

[13] Chen, J. I. Z., & Zong, J. I. (2021). Automatic vehicle license plate detection using K-means clustering algorithm and CNN. Journal of Electrical Engineering and Automation, 3(1), 15-23.

[14] Georganos, S., Grippa, T., Niang Gadiaga, A., Linard, C., Lennert, M., Vanhuysse, S., ... & Kalogirou, S. (2021). Geographical random forests: a spatial extension of the random forest algorithm to address spatial heterogeneity in remote sensing and population modelling. Geocarto International, 36(2), 121-136.

[15] Gupta, V. K., Gupta, A., Kumar, D., & Sardana, A. (2021). Prediction of COVID-19 confirmed, death, and cured cases in India using random forest model. Big Data Mining and Analytics, 4(2), 116-123.

[16] Zhang, W., Wu, C., Zhong, H., Li, Y., & Wang, L. (2021). Prediction of undrained shear strength using extreme gradient boosting and random forest based on Bayesian optimization. Geoscience Frontiers, 12(1), 469-477.

[17] Liu, K., Hu, X., Zhou, H., Tong, L., Widanage, W. D., & Marco, J. (2021). Feature analyses and modeling of lithium-ion battery manufacturing based on random forest classification. IEEE/ASME Transactions on Mechatronics, 26(6), 2944-2955.

[18] Hunter, E. A., Kluck, A. S., Ramon, A. E., Ruff, E., & Dario, J. (2021). The Curvy Ideal Silhouette Scale: measuring cultural differences in the body shape ideals of young US women. Sex Roles, 84(3), 238-251.

[19] Ramalho, E., Peixoto, E., & Medeiros, E. (2021). Silhouette 4D with context selection: Lossless geometry compression of dynamic point clouds. IEEE Signal Processing Letters, 28, 1660-1664.

[20] Shin, J., Chen, F., Lu, C., & Bulut, O. (2022). Analyzing students' performance in computerized formative assessments to optimize teachers' test administration decisions using deep learning frameworks. Journal of Computers in Education, 9(1), 71-91.